\documentclass[11pt]{article}
\usepackage{hyperref}
\usepackage{microtype}
\usepackage[margin=1in]{geometry}
\usepackage[T1]{fontenc}
\usepackage{amsmath,amssymb}
\usepackage{graphicx}
\usepackage{booktabs}
\usepackage{tabularx}
\usepackage{array}
\usepackage{adjustbox}
\usepackage{placeins}
\usepackage{float}
\usepackage[authoryear,round]{natbib}
\usepackage{enumitem}
\setlist{noitemsep,topsep=0.25\baselineskip,leftmargin=*}
\usepackage[nameinlink]{cleveref}

\hypersetup{
  colorlinks=true,
  linkcolor=blue,
  citecolor=blue,
  urlcolor=blue
}

\usepackage{hypernat}

\title{\textbf{Cross-Species Transfer Learning for Electrophysiology-to-Transcriptomics Mapping in Cortical GABAergic Interneurons}}
\author{Theo Schwider, Ramin Ramezani}
\date{}

\begin{document}
\maketitle

\section*{Abstract}
Single-cell electrophysiological recordings provide a powerful window into neuronal functional diversity and offer an interpretable route for linking intrinsic physiology to transcriptomic identity. Here, we replicate and extend the electrophysiology-to-transcriptomics framework introduced by Gouwens et al. (2020) using publicly available Allen Institute Patch-seq datasets from both mouse and human cortex. We focus on GABAergic inhibitory interneurons to target a  subclass structure (Lamp5, Pvalb, Sst, Vip) that is comparable and conserved across species. After quality control, we analyzed 3,699 mouse visual cortex neurons and 506 human neocortical neurons from neurosurgical resections. Using standardized electrophysiological features and sparse PCA, we reproduced the major class-level separations reported in the original mouse study. For supervised prediction, a class-balanced random forest provided a strong feature-engineered baseline in mouse data and a reduced but still informative baseline in human data. We then developed an attention-based BiLSTM that operates directly on the structured IPFX feature-family representation, avoiding sPCA and providing feature-family-level interpretability via learned attention weights. Finally, we evaluated a cross-species transfer setting in which the sequence model is pretrained on mouse data and fine-tuned on human data for an aligned 4-class task, improving human macro-F1 relative to a human-only training baseline. Together, these results confirm reproducibility of the Gouwens pipeline in mouse data, demonstrate that sequence models can match feature-engineered baselines, and show that mouse-to-human transfer learning can provide measurable gains for human subclass prediction. 

\section{Introduction}
Electrophysiological recordings provide a functional view of neuronal identity. Action potential dynamics, firing adaptation, and subthreshold responses reflect the biophysical mechanisms that shape computation in cortical circuits. A central goal in systems and cellular neuroscience is to connect these intrinsic physiological signatures to molecularly defined cell types so that transcriptomic taxonomies can be interpreted in terms of circuit function (Gouwens et al. 2020). Patch-seq has accelerated progress toward this goal by pairing whole-cell recordings with single-cell RNA sequencing (and often morphology) from the same neuron, enabling direct mappings between physiology and transcriptomic identity (Cadwell et al. 2016; Lee et al. 2021; Lipovsek et al. 2021).

Large Patch-seq atlases, particularly from the Allen Institute, make it possible to learn these mappings from thousands of neurons under standardized protocols. In this setting, reproducible feature extraction is essential. Toolchains such as IPFX summarize current-clamp recordings into interpretable electrophysiological feature families (e.g., spike waveform, after hyperpolarization, sag, adaptation, and stimulus-specific summaries), supporting comparability across Neurodata Without Borders (NWB) files and experiments (Allen Institute for Brain Science 2025; Rübel et al.
2022; Teeters et al. 2015). Gouwens et al. showed that such engineered features preserve major transcriptomic separations among inhibitory interneuron subclasses and enable supervised prediction of transcriptomic identity, establishing an influential baseline for multimodal cell-type modeling (Gouwens et al. 2020).

Patch-seq also sits within a broader multimodal literature that links intrinsic physiology with transcriptomics and morphology. Early studies established that combining electrophysiology with single-cell RNA-seq can resolve neuronal subtypes beyond either modality alone (Cadwell et al. 2016; Fuzik et al. 2016), and subsequent work scaled these protocols into high-fidelity pipelines suitable for large atlases (Lee et al. 2021; Lipovsek et al. 2021). Complementary efforts integrated morpho-electric and transcriptomic data across cortical cell types and demonstrated that electrophysiological signatures can predict molecular identity in broader settings (Gouwens et al. 2019; Nandi et al. 2022). At the same time, transcriptomic taxonomies have been shown to generalize across cortical areas and species while retaining meaningful divergence, underscoring both opportunity and challenge for cross-species modeling (Tasic et al. 2018; Hodge et al. 2019).

Extending electrophysiology-to-transcriptomics modeling to human cortex neurons is scientifically and translationally important but introduces additional challenges. Human Patch-seq datasets are typically smaller and more imbalanced, and cross-species comparisons can involve both biological divergence and experimental distribution shift (Hodge et al. 2019; Lee et al. 2023). Recent work has highlighted that some interneuron populations exhibit morphoelectric and transcriptomic divergence across species, motivating approaches that can generalize despite such shifts  (Chartrand et al. 2023). These constraints make transfer learning a natural strategy: mouse data provides abundant labeled examples that may act as auxiliary supervision for limited human datasets.

In this study, we pursue three goals. First, we reproduce the Gouwens et al. baseline on publicly available Allen Institute mouse and human Patch-seq datasets hosted on the DANDI Archive (Gouwens et al. 2020). Second, we develop a sequence model---an attention-based BiLSTM---that operates directly on the structured IPFX feature-family representation, enabling end-to-end learning without sparse PCA and providing interpretability via learned attention over feature families (Bahdanau et al. 2015; Hochreiter and Schmidhuber 1997). Third, we evaluate cross-species transfer by pretraining the sequence model on mouse data and fine-tuning on an aligned human subclass task, testing whether mouse-derived representations improve human subclass prediction under the same evaluation protocol. Together, these analyses provide a replication of a widely used baseline, a deeper characterization of sequence modeling for electrophysiology-derived inputs, and an explicit test of mouse-to-human transfer learning for cell-type prediction.

\section{Background: granular methodological context}

\subsection{Patch-seq modality and cross-species variability}
Patch-seq links intrinsic electrophysiology and transcriptomic identity by measuring both modalities from the same neuron (Gouwens et al. 2020; Cadwell et al. 2016; Lipovsek et al. 2021). Cross-species comparisons additionally contend with systematic differences in tissue source (mouse acute slice vs. human neurosurgical resection), experimental conditions, and protocol coverage, all of which can affect both recorded physiology and downstream annotation quality (Gouwens et al. 2020; Lee et al. 2023). These considerations motivate standardized feature extraction, explicit label harmonization, and conservative train/test separation.

\subsubsection{Cell-type taxonomy and label harmonization}
To enable direct comparisons between mouse and human inhibitory interneurons, we harmonized transcriptomic annotations to a shared inhibitory subclass taxonomy. We mapped fine transcriptomic types (t-types) to one of four canonical GABAergic subclasses---\texttt{Lamp5}, \texttt{Pvalb}, \texttt{Sst}, or \texttt{Vip}---using subclass labels provided in the Allen Institute Patch-seq metadata, guided by cross-species subclass correspondences reported by Hodge et al.\ (2019). For consistency throughout text, tables, and figures, we standardized label capitalization to \texttt{Lamp5}/\texttt{Pvalb}/\texttt{Sst}/\texttt{Vip} (rather than variants such as \texttt{PVALB}, \texttt{PV}, or \texttt{VIP}). Cells annotated as \texttt{Sncg} were treated as \texttt{Vip}-aligned for visualization and downstream analyses, consistent with how \texttt{Sncg}-enriched Vip-associated interneurons are handled in the mouse V1 Patch-seq framework (Gouwens et al., 2020).

\subsection{Engineered electrophysiology features as a reproducible representation}
We adopt the Allen Institute/IPFX representation in which current-clamp recordings are summarized into standardized, interpretable electrophysiological feature families (Gouwens et al. (2020); Allen Institute for Brain Science 2025). This representation supports scale (thousands of cells), increases comparability across NWB files, and yields features with clear mechanistic interpretations (e.g., spike waveform, adaptation, and subthreshold properties) that are useful both for embedding and for supervised prediction.

\subsection{Dimensionality Reduction}
Sparse PCA (sPCA) provides a low-dimensional representation while keeping components relatively interpretable via sparse loadings (Zou et al. 2006). UMAP then provides a nonlinear embedding used here mainly for qualitative validation and visualization of class-level separations (McInnes et al. 2018). In this study, these methods are used as analysis tools rather than as an end-to-end optimized preprocessing pipeline.

\subsection{Class imbalance and metric choice}
Inhibitory subclasses are unevenly sampled, particularly in the smaller human dataset. For that reason, we emphasize class-balanced evaluation (macro-F1 and balanced accuracy) so that performance is not dominated by majority classes (Breiman 2001). When used, SMOTE oversampling is applied only to the training partition to mitigate imbalance without contaminating held-out evaluation (Chawla et al. 2002).

\section{Materials and Methods}
\subsection{Data sources}
We analyzed patch-clamp electrophysiological recordings from both mouse and human cortex obtained from publicly available datasets released by the Allen Institute for Brain Science and hosted on the DANDI Archive (Allen Institute for Brain Science, 2021, 2024). All analyses were restricted to GABAergic inhibitory interneurons (human inhibitory subclasses; mouse interneuron subclasses), yielding aligned subclass labels for cross-species experiments. Mouse data was drawn from the visual cortex Patch-seq characterization project (DANDI:000020) (Allen Institute for Brain Science, 2021), while human data was taken from Patch-seq recordings of neurosurgical cortex resections (DANDI:000636) (Allen Institute for Brain Science, 2024). Human metadata/annotations used in this work were taken from the Allen Institute human\_patchseq\_gaba repository associated with the Lee et al. (2023) study (Allen Institute for Brain Science, 2023; Lee et al., 2023). For each dataset, we downloaded Neurodata Without Borders (NWB) files containing raw current-clamp recordings, stimulus annotations, and cell-level metadata (Teeters et al., 2015; R\"ubel et al., 2022).

In total, we analyzed 3,699 cells from the mouse visual cortex and 506 cells from the human neocortex after quality control (see below) (Allen Institute for Brain Science, 2021, 2024). For all analyses, we restricted our attention to neurons with at least one valid step-current injection protocol suitable for standard electrophysiological feature extraction (Gouwens et al., 2020).

\subsection{Dataset variants and splits}
Table~\ref{tab:dataset-variants} summarizes the dataset variants used throughout the study and clarifies the label spaces, split strategies, and any train-time augmentation. Mouse experiments were run both in the original 5-class label space (including \texttt{Sncg}) and in an aligned 4-class label space in which \texttt{Sncg} was merged into \texttt{Vip} to match the human annotation granularity. Human experiments were performed in the 4-class space and evaluated using stratified cross-validation, with any oversampling applied to training data only.
\begin{table}[t]
\centering
\small
\setlength{\tabcolsep}{4pt}
\renewcommand{\arraystretch}{1.15}
\begin{adjustbox}{max width=\linewidth}
\begin{tabularx}{\linewidth}{@{}>{\raggedright\arraybackslash}p{0.22\linewidth} >{\raggedright\arraybackslash}p{0.18\linewidth} r >{\raggedright\arraybackslash}X >{\raggedright\arraybackslash}p{0.12\linewidth} >{\raggedright\arraybackslash}p{0.14\linewidth}@{}}
\toprule
\textbf{Dataset / variant} & \textbf{Source} & \textbf{$N$} & \textbf{Label space} & \textbf{Split} & \textbf{Train aug.} \\
\midrule
Mouse (base) & DANDI:000020 (Allen Institute for Brain Science, 2021) & 3699 & 5-class (Lamp5, Pvalb, Sst, Vip, Sncg) & 60/20/20 & None \\
Mouse (aligned) & DANDI:000020 (Allen Institute for Brain Science, 2021) & 3699 & 4-class (Lamp5, Pvalb, Sst, Vip; Vip = Vip+Sncg) & 60/20/20 & None \\
Mouse + SMOTE (train-only) & DANDI:000020 (Allen Institute for Brain Science, 2021) & 3699 & 5-class (Lamp5, Pvalb, Sst, Vip, Sncg) & 60/20/20 & SMOTE (train only) \\
Human + SMOTE (train-only) & DANDI:000636 + \mbox{human\_patchseq\_gaba} (Allen Institute for Brain Science, 2024, 2023) & 506 & 4-class (Lamp5, Pvalb, Sst, Vip) & 60/20/20 & SMOTE (train only) \\
Dual / joint (mouse + human) & Mouse: DANDI:000020; Human: DANDI:000636 + \mbox{human\_patchseq\_gaba} (Allen Institute for Brain Science, 2021, 2024, 2023) & 3699+506 & Shared encoder + two heads; reported on aligned 4-class & per species & Mixed + fine-tune \\
\bottomrule
\end{tabularx}
\end{adjustbox}
\caption{Dataset variants used in this study, including label-space alignment and train-time augmentation. Mouse Patch-seq data are from DANDI:000020 and human Patch-seq data are from DANDI:000636, with additional human inhibitory subclass metadata from the Allen Institute \texttt{human\_patchseq\_gaba} repository associated with Lee et al.\ (2023).}
\label{tab:dataset-variants}
\end{table}

\subsection{Inclusion criteria and quality control}
Cells were included if the NWB file contained at least one current-clamp step stimulus with clearly annotated sweep numbers and stimulus amplitudes (Teeters et al., 2015; R\"ubel et al., 2022), the recording passed the basic quality checks implemented in the Allen Institute IPFX pipeline (e.g., stable baseline, absence of severe artifacts, and a minimum recording duration sufficient to capture stimulus onset and offset) (Allen Institute for Brain Science, 2025; Gouwens et al., 2020), and the metadata contained a labeled transcriptomic type (Gouwens et al., 2020; Lee et al., 2023). For feature extraction, we further required that each cell exhibited either at least one action potential in response to a depolarizing step (to support spike-based feature families) or a clean subthreshold response (to support subthreshold feature families) (Allen Institute for Brain Science, 2025; Gouwens et al., 2020).\sloppy

Sweeps that failed IPFX quality-control checks (e.g., missing stimulus metadata, excessive noise, unstable baseline) were excluded on a per-sweep basis. Cells with no remaining usable sweeps were removed from further analysis.

\subsection{Electrophysiological feature extraction}

\makeatletter
\renewcommand{\NAT@citesuper}[3]{\hyper@natlinkstart{#1}\textsuperscript{#2}\hyper@natlinkend}
\makeatother
Electrophysiological features were extracted using the IPFX Python package (Allen Institute) following the approach used in Gouwens et al. (2020) (Allen Institute for Brain Science, 2025; Gouwens et al., 2020). For each cell, we identified the set of step-current sweeps and applied the standard feature extraction routines to compute a panel of biophysically interpretable features.

We organized features into 12 families analogous to the Gouwens feature taxonomy displayed in Table~\ref{tab:ipfx-feature-families}.
\begin{table}[H]
\centering
\renewcommand{\arraystretch}{1.25}
\setlength{\tabcolsep}{10pt}
\small
\begin{tabularx}{\linewidth}{@{}>{\ttfamily}p{0.34\linewidth}X@{}}
\toprule
\normalfont\textbf{Feature family} & \textbf{Description} \\
\midrule
\addlinespace[2pt]
first\_ap\_v & Voltage features of the first action potential, such as threshold, peak, amplitude, and afterhyperpolarization. \\
\addlinespace[2pt]
first\_ap\_dv & Derivatives of the membrane potential around the first spike (e.g., $dV/dt$ at threshold and peak). \\
\addlinespace[2pt]
isi\_shape & Inter-spike interval metrics, including adaptation indices and ISI variability. \\
\addlinespace[2pt]
inst\_freq & Instantaneous firing rate measures as a function of time and stimulus amplitude. \\
\addlinespace[2pt]
spiking\_threshold\_v & Spike threshold estimates across sweeps. \\
\addlinespace[2pt]
spiking\_peak\_v & Spike peak voltages across sweeps. \\
\addlinespace[2pt]
spiking\_width & Spike half-width and related temporal width measures. \\
\addlinespace[2pt]
spiking\_fast\_trough\_v & Post-spike fast trough amplitudes. \\
\addlinespace[2pt]
spiking\_upstroke\_downstroke\_ratio & Ratios of maximum upstroke and downstroke derivatives. \\
\addlinespace[2pt]
step\_subthresh & Subthreshold voltage responses to step stimuli (e.g., sag and related measures). \\
\addlinespace[2pt]
subthresh\_norm & Normalized subthreshold response features, including input resistance and membrane time constant estimates. \\
\addlinespace[2pt]
psth & Peristimulus time histogram--like summaries of spiking over time for a given step stimulus. \\
\addlinespace[2pt]
\bottomrule
\end{tabularx}
\normalsize
\renewcommand{\arraystretch}{1.0}
\caption{Summary of the 12 IPFX electrophysiological feature families used in this study, following the feature-family taxonomy described by Gouwens et al.\ (2020) and extracted using the Allen Institute IPFX pipeline (Allen Institute for Brain Science, 2025; Gouwens et al., 2020).}
\label{tab:ipfx-feature-families}
\end{table}

When multiple sweeps of the same type were available, IPFX summary statistics (e.g., mean or median across sweeps) were used. For feature families that could not be computed for a given cell (e.g., non-spiking cells lacking spike-based features), we initially set corresponding values to NaN. Cells with excessive missingness were removed; for the remaining cells, residual missing values were imputed using feature-wise medians computed separately within each species.

\subsection{Feature preprocessing}
All feature vectors were assembled into a fixed-order representation of shape $(N_{\text{cells}}, N_{\text{families}}, N_{\text{features per family}})$, where $N_{\text{families}} = 12$. To ensure consistent ordering across datasets and experiments, we defined a canonical family ordering (matching the list above) and stored it alongside the feature matrices.

Prior to model training, we applied three preprocessing steps. First, we log-transformed strictly positive features with highly skewed distributions (e.g., firing rates). Second, we performed within-species Z-score normalization using statistics computed on the training set only, and then applied the same transformation to validation and test sets. Third, for neural network models, we reshaped the features into sequences of length 12 (feature families) with a feature dimension corresponding to the concatenated features within each family, yielding an input shape (sequence\_length=12, feature\_dim=498).

\subsection{Machine-learning models}
\subsubsection{Random forest classifier (baseline)}
As a simple, interpretable baseline, we trained a random forest classifier using scikit-learn (Breiman, 2001). Before using the random forest we followed Gouwens’ procedure and computed sparse PCA to featurize each feature family, yielding 44 sparse PCs total (Zou et al., 2006; Gouwens et al., 2020). Trees were trained on the flattened sparse component vectors. For evaluation, we performed 10 independent stratified train/test splits and reported the mean of accuracy, macro-F1, and balanced accuracy across runs. In each run, we trained a \texttt{RandomForestClassifier} with 600 trees, bootstrap sampling enabled, \texttt{min\_samples\_leaf} = 2, and \texttt{class\_weight} = \texttt{balanced\_subsample}. Trees were grown to full depth (\texttt{max\_depth} = \texttt{None}) with default splitting rules (\texttt{min\_samples\_split} = 2). This model served as an interpretable baseline for both mouse and human datasets, using the same accuracy-testing procedure as the LSTM counterparts.

\subsubsection{LSTM neural network}
For the main experiments, we used a recurrent neural network to exploit the structured organization of electrophysiological feature families. Inputs were encoded as sequences of length 12 (one step per feature family) with per-step feature dimension. All models shared a common PyTorch backbone and were improved in stages:
The baseline model was a single-layer bidirectional LSTM with hidden size $H=128$ per direction; its hidden states were aggregated into a fixed-length representation for classification (Hochreiter \& Schmidhuber, 1997). We then added a self-attention aggregation over the 12 hidden states to learn feature-family importance and to form a weighted context vector (Bahdanau et al., 2015). To address class imbalance, we optionally applied SMOTE oversampling on the training split only (never on validation/test) while keeping the BiLSTM+attention architecture unchanged (Chawla et al., 2002). Finally, we evaluated an ArcFace-style angular-margin classifier head in place of the standard softmax classifier to improve embedding separability and robustness, especially for minority classes (Deng et al., 2019); when using class-balanced loss variants we referenced the effective-number-of-samples formulation (Cui et al., 2019), and when using focal-style reweighting we referenced focal loss (Lin et al., 2017).

The resulting progression---BiLSTM $\to$ BiLSTM+Attention $\to$ BiLSTM+Attention+SMOTE $\to$ ArcFace BiLSTM+Attention+SMOTE---captures each incremental improvement evaluated in our mouse experiments.

We trained the networks using categorical cross-entropy loss and the Adam optimizer (initial learning rate $1\times 10^{-3}$, $\beta_1 = 0.9$, $\beta_2 = 0.999$). Mini-batch size was set to 64. Models were trained for up to 50 epochs with early stopping based on validation macro-averaged F1 score, with a patience of 7 epochs.

\subsubsection{Attention-weight analysis (interpretability)}
For our trained LSTM+attention models, we extracted the learned attention weights over the 12 feature-family timesteps (softmax-normalized across families for each cell) (Bahdanau et al., 2015). We then computed mean attention weights per family for each transcriptomic class, and aggregated these means across runs by weighting each run’s class mean by the number of samples contributing to that class. This analysis does not establish causality, but provides an interpretable summary of which feature families the model emphasizes when forming its pooled representation.

\subsection{Transfer learning configurations}
We initially began with a sequential transfer setup: pretrain the encoder on the mouse aligned dataset, then fine-tune on human using a smaller human train/validation split. In practice, this regime was unstable across runs, especially for minority human classes, due to distribution shift, imbalance interactions, and limited validation signal.

To reduce this sensitivity, we introduced joint supervised training so that human supervision shapes the encoder from the beginning, while mouse data provides a stabilizing auxiliary signal. The two-head design prevents label-space conflicts: mouse gradients improve shared feature extraction without forcing a single classifier to reconcile cross-species label mismatches. Finally, the human-only fine-tuning stage ensures the final model is explicitly adapted to the human decision boundary before evaluation on the held-out human test set.

\subsubsection{Joint supervised training (shared encoder, two heads)}
Joint supervised training used one shared encoder (input adapter, normalization, BiLSTM, and attention) that feeds two separate classifiers, one predicting mouse labels and the other predicting human labels. Training used mixed mini-batches from mouse and human in each epoch and optimized a joint objective consisting of the human loss plus a weighted mouse loss, where $\alpha$ controls the mouse contribution. Model selection used early stopping based on human validation macro-F1.

\subsubsection{Human-only fine-tuning after joint training}
After joint training, we initialized the human model by copying the shared encoder weights and the human head weights from the joint model. We then continued training on human data only using the same human train/validation split and reported performance on the held-out human test set.

\subsection{Train/validation/test splits and class imbalance}
\textbf{Mouse experiments (hold-out split).} For mouse experiments, cells were split into training, validation, and test sets using stratified sampling by class (60\%/20\%/20\%). All hyperparameter tuning and early stopping decisions were made exclusively on the validation set. The held-out test set was used only for final performance reporting.

\textbf{Human experiments (5-fold stratified cross-validation).} For human experiments, we used 5-fold stratified cross-validation at the cell level. In each fold, we trained on 4 folds and evaluated on the remaining fold; we report the mean and standard deviation across folds. All oversampling (SMOTE) and all normalization statistics were computed using the training partition of each fold only.

\textbf{Class imbalance} Class imbalance was addressed using a combination of techniques:
We used class weighting in the loss function and, when needed, oversampled minority classes with SMOTE; SMOTE was applied to the training data only to avoid leakage into validation/test partitions (mouse) or held-out folds (human).

\subsection{Evaluation metrics and statistical analysis}
Model performance was primarily assessed using macro-averaged F1 score and accuracy, which gives equal weight to each class irrespective of its frequency. We also report class-wise precision and recall, and confusion matrices for selected models. For multi-run experiments, each configuration was trained with 10 different random seeds. Human results are reported as mean $\pm$ standard deviation across the 5 stratified folds.

\section{Results}
\subsection{Dataset composition and feature coverage}
After applying the inclusion criteria and sweep-level quality control, we retained 3,699 mouse neurons from primary visual cortex and 506 human neurons from neurosurgical neocortical resections for analysis. Across both species, the majority of cells had at least one depolarizing step that elicited action potentials, enabling computation of the full set of spike-based feature families. Following removal of cells with extensive missingness and median imputation of residual NaNs within species, the final feature tensor contained complete 12-family profiles for essentially all retained neurons, with comparable coverage across mouse and human datasets. These retained sample sizes and subclass counts (mouse 5-class; human 4-class) are summarized in Table~\ref{tab:class-distributions}.

\begin{table}[t]
\centering
\small
\begin{tabular}{@{}lr@{}}
\toprule
\textbf{Mouse data class} & \textbf{Count (\%)} \\
\midrule
Lamp5 & 402 (10.9\%) \\
Pvalb & 745 (20.1\%) \\
Sncg & 198 (5.4\%) \\
Sst & 1663 (45.0\%) \\
Vip & 691 (18.7\%) \\
\bottomrule
\end{tabular}
\hspace{0.75em}
\begin{tabular}{@{}lr@{}}
\toprule
\textbf{Human data class} & \textbf{Count (\%)} \\
\midrule
Lamp5 & 50 (9.9\%) \\
Pvalb & 293 (57.9\%) \\
Sst & 96 (19.0\%) \\
Vip & 67 (13.2\%) \\
\bottomrule
\end{tabular}
\caption{Class distributions after QC (as reported in the draft).}
\label{tab:class-distributions}
\end{table}

\subsection{Baseline random forest performance on mice and humans}
\FloatBarrier
\begin{figure}[!htbp]
\centering
\begin{minipage}[t]{0.49\linewidth}
  \centering
  \includegraphics[width=\linewidth,height=0.32\textheight,keepaspectratio]{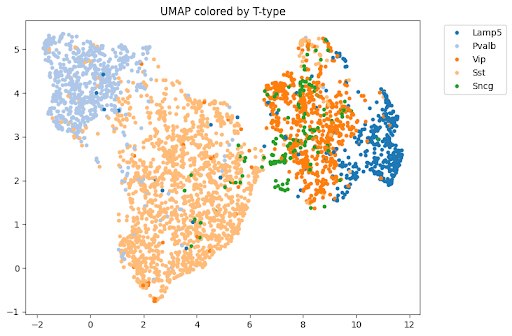}
  \vspace{0.25em}
  \textbf{(A)} Our UMAP
\end{minipage}\hfill
\begin{minipage}[t]{0.49\linewidth}
  \centering
  \includegraphics[width=\linewidth,height=0.32\textheight,keepaspectratio]{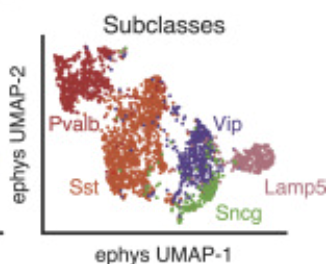}
  \vspace{0.25em}
  \textbf{(B)} Gouwens baseline UMAP \\ (Gouwens et al. 2020)
\end{minipage}
\caption{UMAP comparison used to validate replication of the baseline feature pipeline.}
\label{fig:umap-comparison}
\end{figure}

To reproduce the Gouwens et al.\ (2020) baseline pipeline, we used sparse PCA (sPCA) to reduce the IPFX-derived electrophysiological feature panel into 44 sparse principal components (sPCs; each explaining at least 1\% of variance), and embedded cells with UMAP for visualization. The resulting mouse UMAP reproduced the major class-level separations reported in the original study (Pvalb, Sst, Lamp5), while Vip and Sncg exhibited the largest overlap (Figure~\ref{fig:umap-comparison}).

\begin{figure}[H]
\centering
\includegraphics[width=0.85\linewidth]{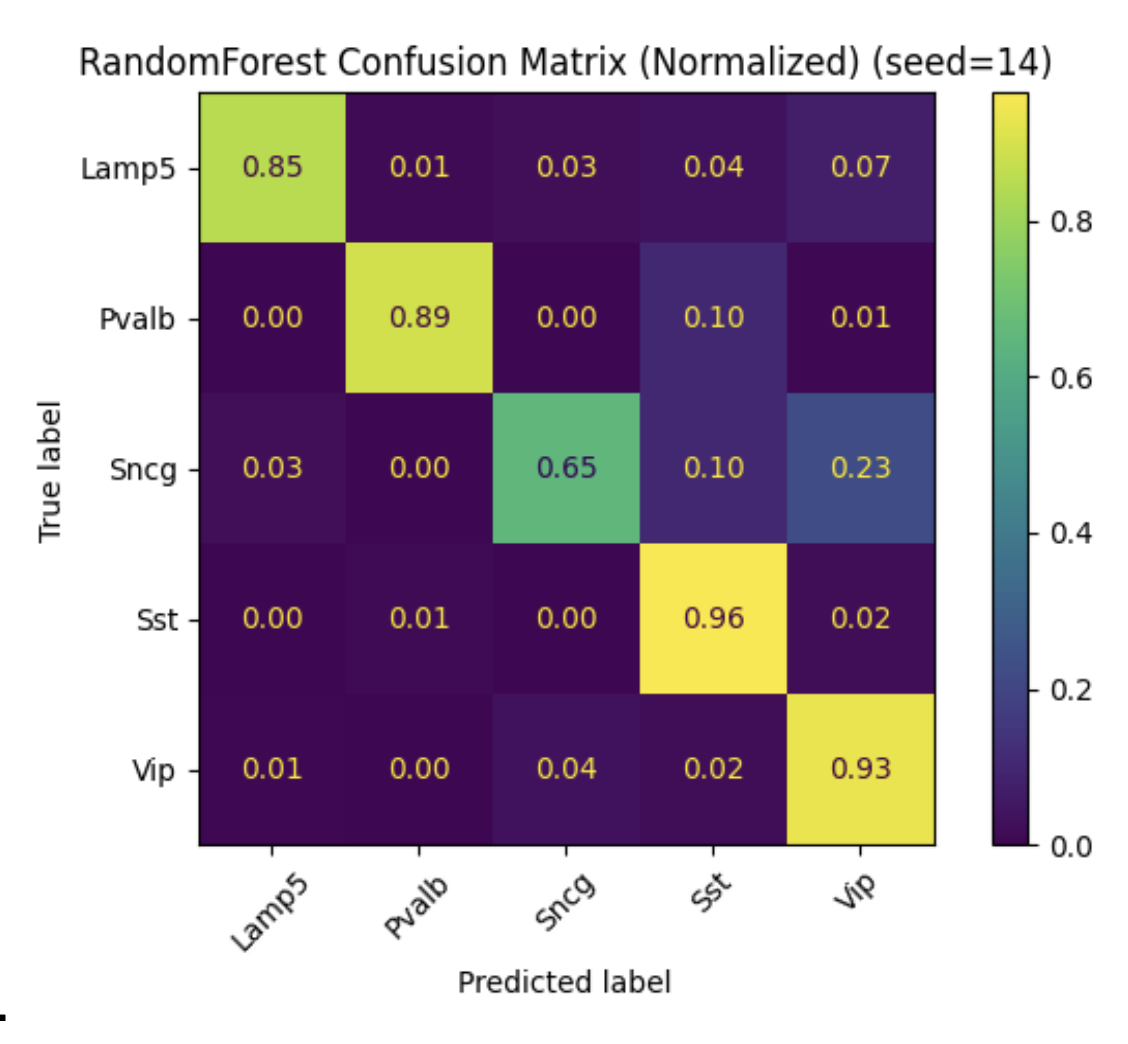}
\caption{Random forest confusion matrix on mouse test split (5-class).}
\label{fig:rf-mouse-confusion}
\end{figure}

For supervised prediction, we trained a class-balanced random forest classifier using the 44 sPCs as input features. On the mouse held-out test split, the model achieved 90.72\% accuracy and 0.8728 macro-F1 (mean over 10 random seeds; Figure~\ref{fig:rf-mouse-confusion}).

\begin{figure}[H]
\centering
\includegraphics[width=0.85\linewidth]{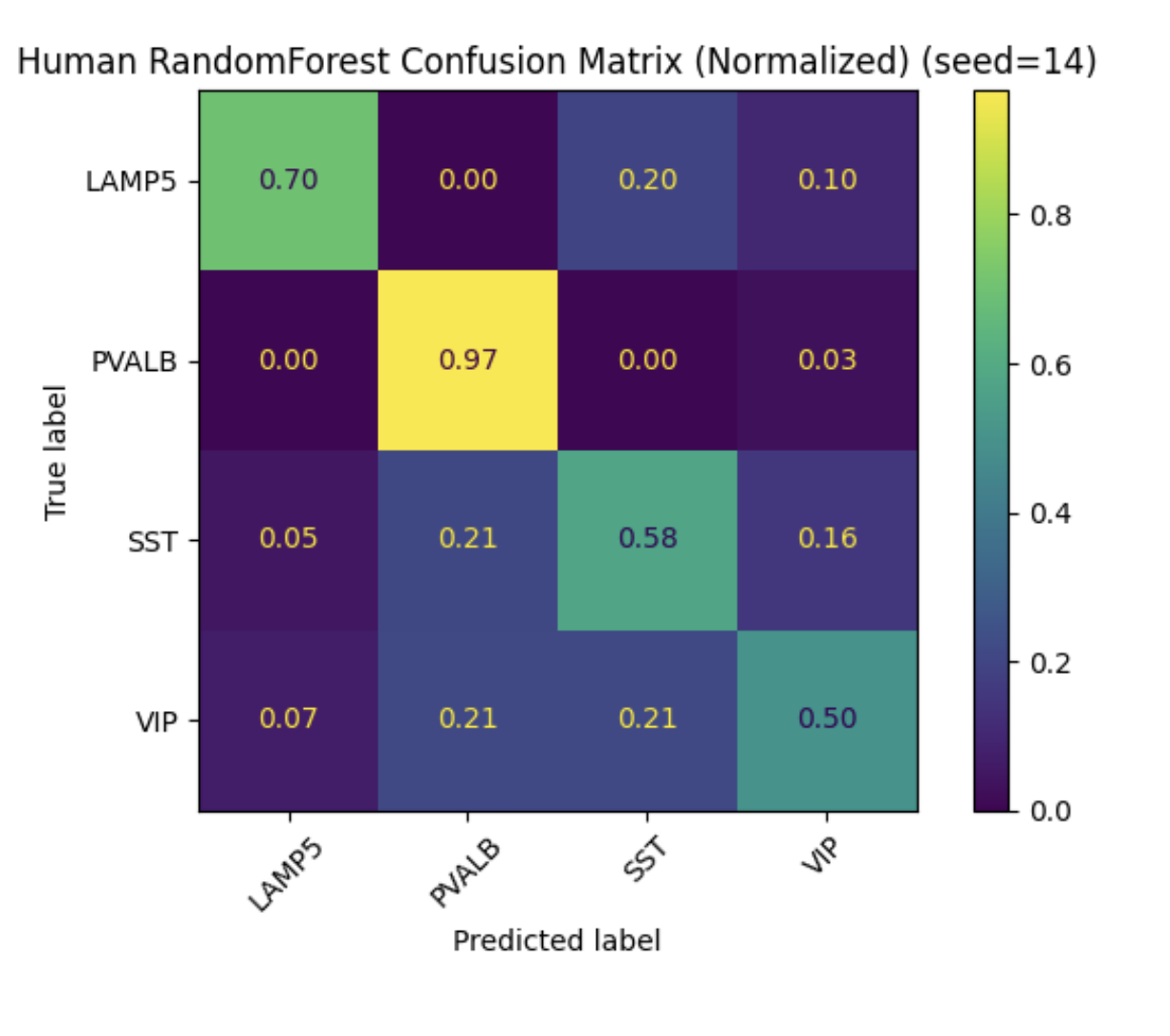}
\caption{Random forest confusion matrix on human evaluation (4-class; 5-fold stratified cross-validation).}
\label{fig:rf-human-confusion}
\end{figure}

Applying the same baseline procedure to the human dataset (sPCA followed by a random forest, evaluated with 5-fold stratified cross-validation) yielded 75.18\% accuracy and 0.6589 macro-F1 on average (Figure~\ref{fig:rf-human-confusion}), consistent with reduced sample size and stronger class imbalance in the human data.

The same baseline is predictive in human but degrades with sample size/imbalance. Applying the same procedure to the human dataset (sPCA then random forest, evaluated with 5-fold stratified cross-validation) yielded 75.18\% accuracy and 0.6589 macro-F1 on average (Figure~\ref{fig:rf-human-confusion}).

\FloatBarrier

\clearpage
\subsection{LSTM performance on mouse and human datasets}
\begin{table}[H]
\centering
\small
\begin{adjustbox}{max width=\linewidth}
\begin{tabular}{@{}lcc@{}}
\toprule
\textbf{Mouse model variant} & \textbf{Macro-F1 (avg over 10 runs)} & \textbf{Accuracy (avg over 10 runs)} \\
\midrule
Baseline BiLSTM & 0.8601 $\pm$ 0.0162 & 0.9062 $\pm$ 0.0091 \\
BiLSTM with attention mechanism & 0.8738 $\pm$ 0.0136 & 0.9145 $\pm$ 0.0088 \\
BiLSTM with attention mechanism and SMOTE & 0.8856 $\pm$ 0.0214 & 0.9193 $\pm$ 0.0128 \\
ArcFace BiLSTM with attention mechanism and SMOTE & 0.8923 $\pm$ 0.0175 & 0.9235 $\pm$ 0.0093 \\
\bottomrule
\end{tabular}
\end{adjustbox}
\caption{Mouse LSTM performance (as reported in the draft).}
\label{tab:lstm-mouse}
\end{table}

\begin{table}[H]
\centering
\small
\begin{adjustbox}{max width=\linewidth}
\begin{tabular}{@{}lcc@{}}
\toprule
\textbf{Human model variant} & \textbf{Macro-F1 (5-fold avg over 10 runs)} & \textbf{Accuracy (5-fold avg over 10 runs)} \\
\midrule
BiLSTM with attention mechanism & 0.6685 $\pm$ 0.0141 & 0.7798 $\pm$ 0.0080 \\
BiLSTM with attention mechanism and SMOTE & 0.6754 $\pm$ 0.0180 & 0.7822 $\pm$ 0.0145 \\
ArcFace BiLSTM with attention mechanism and SMOTE & 0.6729 $\pm$ 0.0195 & 0.7818 $\pm$ 0.0145 \\
\bottomrule
\end{tabular}
\end{adjustbox}
\caption{Human LSTM performance (as reported in the draft).}
\label{tab:lstm-human}
\end{table}

\FloatBarrier
Treating the 12 IPFX feature families as an ordered sequence, LSTM-based models achieved strong performance without requiring sPCA preprocessing.

\textbf{Mouse.} Across 10 random seeds, macro-F1 improved from 0.8601 (baseline BiLSTM) to 0.8923 (ArcFace BiLSTM+attention+SMOTE), with a corresponding accuracy increase from 0.9062 to 0.9235 (Table~\ref{tab:lstm-mouse}).

\textbf{Human.} Under 5-fold stratified cross-validation, the BiLSTM+attention model achieved 0.6685 macro-F1 and 0.7798 accuracy on average; SMOTE and ArcFace produced only modest changes in these aggregate metrics (Table~\ref{tab:lstm-human}).

\subsection{Attention-based interpretability}
\begin{figure}[H]
\centering
\includegraphics[width=\linewidth]{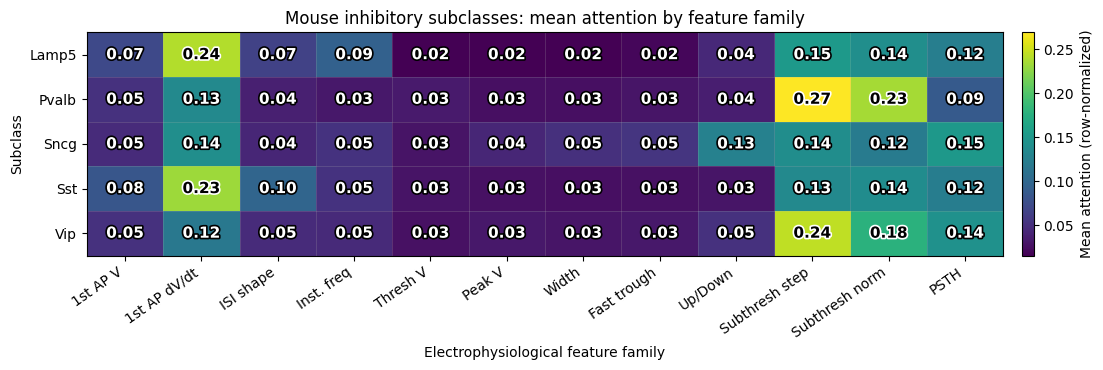}
\caption{Attention-weight summaries for interpretability in mouse (mean attention weights by feature family and class).}
\label{fig:attention-weights-mouse}
\end{figure}

\begin{figure}[H]
\centering
\includegraphics[width=\linewidth]{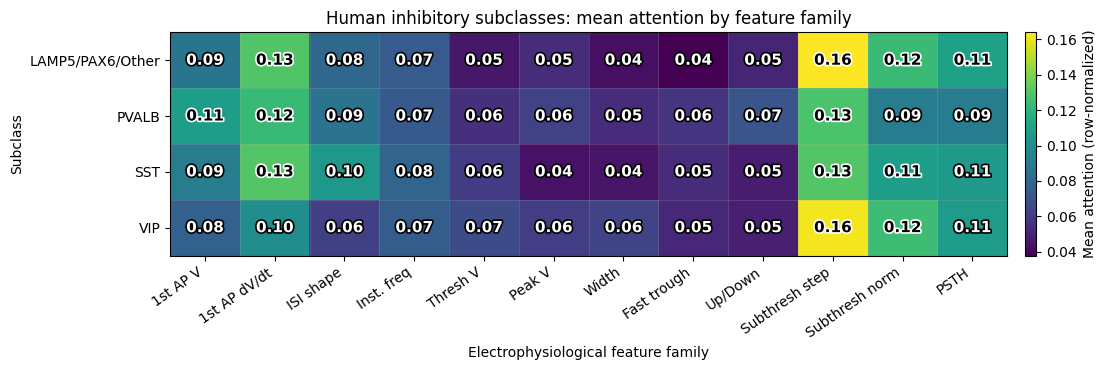}
\caption{Attention-weight summaries for interpretability in human (mean attention weights by feature family and class).}
\label{fig:attention-weights-human}
\end{figure}

To provide an interpretable check on what the models use, we summarized the attention weights assigned to each of the 12 electrophysiological feature families (Figures~\ref{fig:attention-weights-mouse} and \ref{fig:attention-weights-human}). Aggregated across runs, the attention profiles were stable and subclass-specific. In particular for mice, Lamp5 and Sst placed the largest weight on the \texttt{first\_ap\_dv} family, whereas Pvalb emphasized subthreshold-related families (\texttt{step\_subthresh} and \texttt{subthresh\_norm}). Vip and Sncg exhibited more distributed profiles.

\FloatBarrier

\subsection{Effects of transfer learning from mouse to human}
\FloatBarrier
\begin{table}[H]
\centering
\small
\begin{tabular}{@{}lcc@{}}
\toprule
\textbf{Model variant} & \textbf{Macro-F1 (k-fold avg)} & \textbf{Accuracy (k-fold avg)} \\
\midrule
Baseline BiLSTM & 0.6580 $\pm$ 0.0286 & 0.7710 $\pm$ 0.0172 \\
Dual pretrained transfer learning & 0.6795 $\pm$ 0.0120 & 0.7905 $\pm$ 0.0096 \\
\bottomrule
\end{tabular}
\caption{Transfer-learning results on the aligned 4-class task (human evaluated with 5-fold stratified cross-validation).}
\label{tab:transfer}
\end{table}

We evaluated mouse$\to$human transfer on the aligned 4-class inhibitory subclass task by comparing a human-only paired baseline to a mouse+human pretrained, human--fine-tuned model under the same evaluation protocol. As summarized in Table~\ref{tab:transfer}, transfer learning produced a consistent improvement over the baseline: macro-F1 increased from 0.6580 $\pm$ 0.0286 to 0.6795 $\pm$ 0.0120, and accuracy increased from 0.7710 $\pm$ 0.0172 to 0.7905 $\pm$ 0.0096.
\FloatBarrier

\subsection{Class imbalance, error patterns, and robustness}
Across all configurations, class imbalance strongly influenced error patterns. Frequent classes were typically classified with high precision and recall, whereas rare t-types (e.g., Sncg) and sparsely sampled human subclasses exhibited lower macro-F1 and more variable performance across seeds.

\section{Discussion}
In this work, we replicated the baseline electrophysiology-to-transcriptomics pipeline of Gouwens et al.\ (2020) on Allen Institute Patch-seq datasets and extended it to human inhibitory interneurons. Electrophysiology-derived features were strongly predictive of transcriptomic subclass identity in mouse, and remained predictive (though with reduced performance) in human, consistent with the smaller and more imbalanced human dataset. This sample-size limitation is reflected in the subclass distributions (Table~\ref{tab:class-distributions}), which show both the smaller overall human cohort and stronger human subclass imbalance relative to mouse.

The UMAP embeddings and random-forest baseline confirm that engineered IPFX features preserve major subclass-level structure, with particularly clear separations among \texttt{Pvalb}, \texttt{Sst}, and \texttt{Lamp5}. Beyond this feature-engineered baseline, we show that sequence models can operate directly on the structured 12-family IPFX representation. The attention-based BiLSTM variants achieved comparable performance without requiring sparse PCA preprocessing and additionally provided a transparent summary of which feature families contributed most to subclass discrimination.

A key result is that cross-species transfer learning can improve performance on the aligned human 4-class task. Pretraining on mouse data and fine-tuning on human data increased macro-F1 relative to a human-only baseline under the same evaluation protocol, supporting the premise that mouse Patch-seq can provide useful auxiliary supervision when human datasets are small.

Operationally, this improvement comes from extending the original Gouwens-style workflow into a transfer-ready pipeline: standardized NWB/IPFX feature extraction and quality control, cross-species subclass harmonization to the shared \texttt{Lamp5}/\texttt{Pvalb}/\texttt{Sst}/\texttt{Vip} label space, family-structured feature packaging for sequence modeling that lets the BiLSTM/LSTMNN operate directly on the native 12-family IPFX representation (without the sPCA compression step used in the original Gouwens pipeline), mouse pretraining, and human fine-tuning under matched evaluation splits. Framed this way, the pipeline is directly reusable for human transcriptomic subclass classification in data-limited settings: the mouse-trained representation acts as a biologically informed prior, and the human fine-tuning stage adapts that prior to cohort-specific distributions, yielding better macro-F1 and more stable performance than training only on the smaller human set.

The remaining performance gap between mouse and human likely reflects a combination of reduced sample size, stronger class imbalance, and distribution shift due to both biological divergence and experimental differences (e.g., tissue source and recording conditions) (Hodge et al., 2019; Lee et al., 2023; Chartrand et al., 2023). These factors motivate future work that disentangles label noise from true species differences and that evaluates domain-adaptation strategies more explicitly, for example by conditioning on recording metadata, learning species-invariant representations, or incorporating additional modalities when available.

The attention-weight summaries provide an interpretability layer on top of predictive performance. While attention should not be over-interpreted as causal attribution, the resulting class-level profiles were stable across runs and broadly consistent with the notion that different subclasses rely on distinct combinations of spike-shape and subthreshold feature families.

Several limitations remain. First, our conclusions are constrained by dataset scope and label alignment: the mouse sample is large relative to humans, and human Patch-seq labels are inherently noisier and less uniformly sampled across subclasses. Second, our approach relies on engineered features and their summaries; subtle waveform-level differences may be compressed away. Third, while the improved BiLSTM architectures help, we did not exhaustively search model families; more expressive approaches (e.g., transformer-style sequence models or direct voltage-trace ingestion with 1D CNN/transformers) could further reduce fine-grained confusion. Finally, cross-species differences in tissue source, experimental conditions, and biological composition may impose ceilings on the degree to which a mouse-derived feature space can generalize to humans without explicit domain adaptation.

\begin{figure}[H]
\centering
\includegraphics[width=0.72\linewidth]{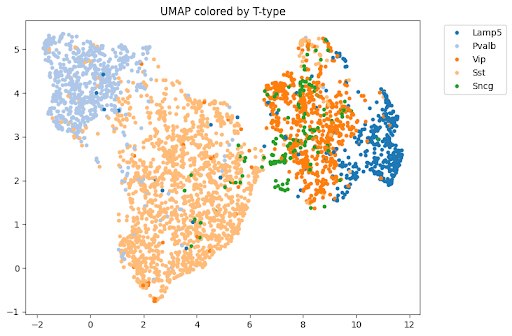}
\caption{Recap figure (reproduced for convenience). UMAP embedding of mouse inhibitory interneurons in the electrophysiology-derived low-dimensional space (IPFX features with the same dimensionality-reduction pipeline described in Materials and Methods). This panel is repeated here as a compact visual reference for the Discussion and does not present additional analyses.}
\label{fig:umap-summary}
\end{figure}

\begin{figure}[H]
\centering
\begin{adjustbox}{max width=\linewidth}
\begin{tabular}{cc}
\includegraphics[width=0.48\linewidth]{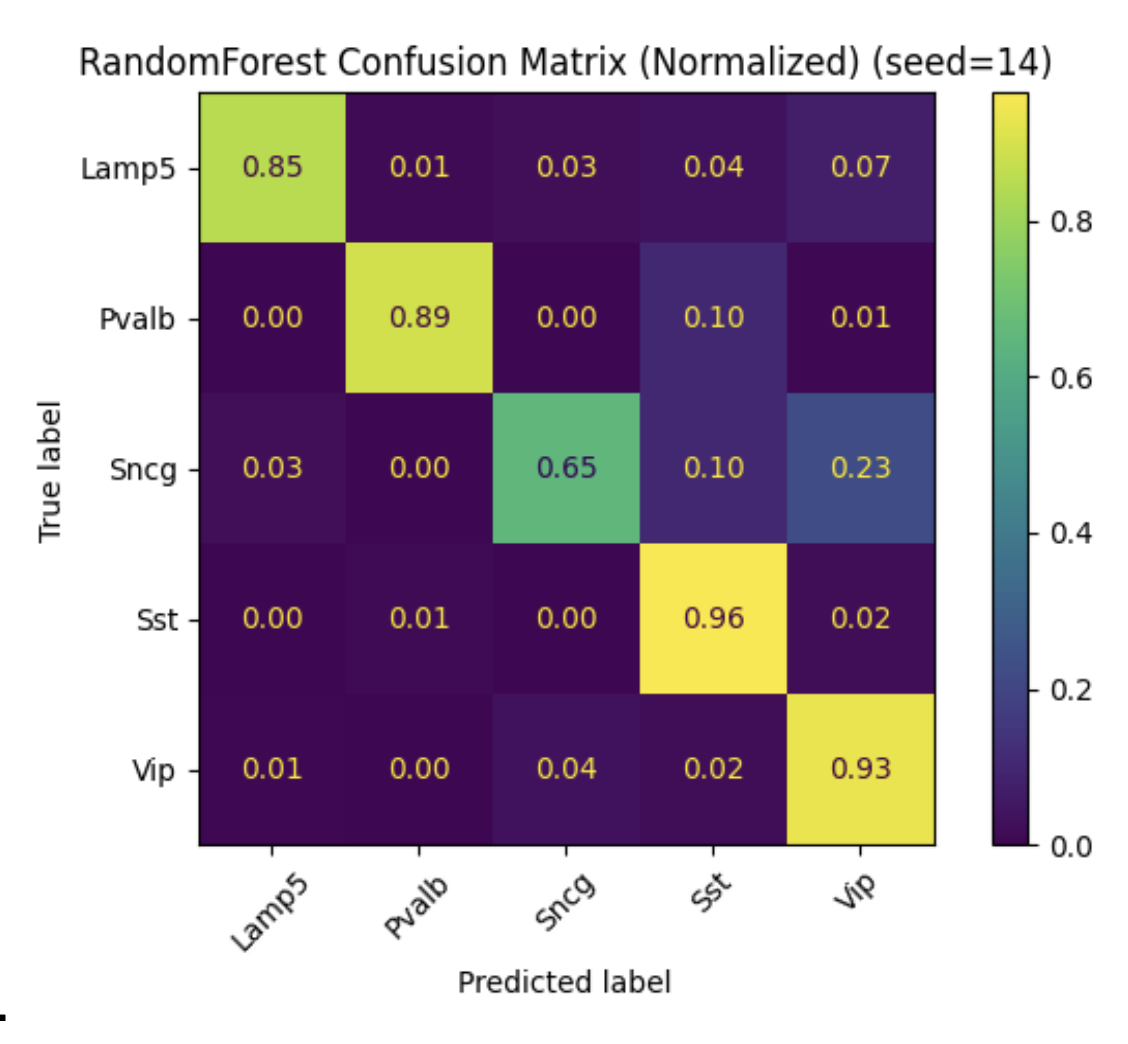} &
\includegraphics[width=0.48\linewidth]{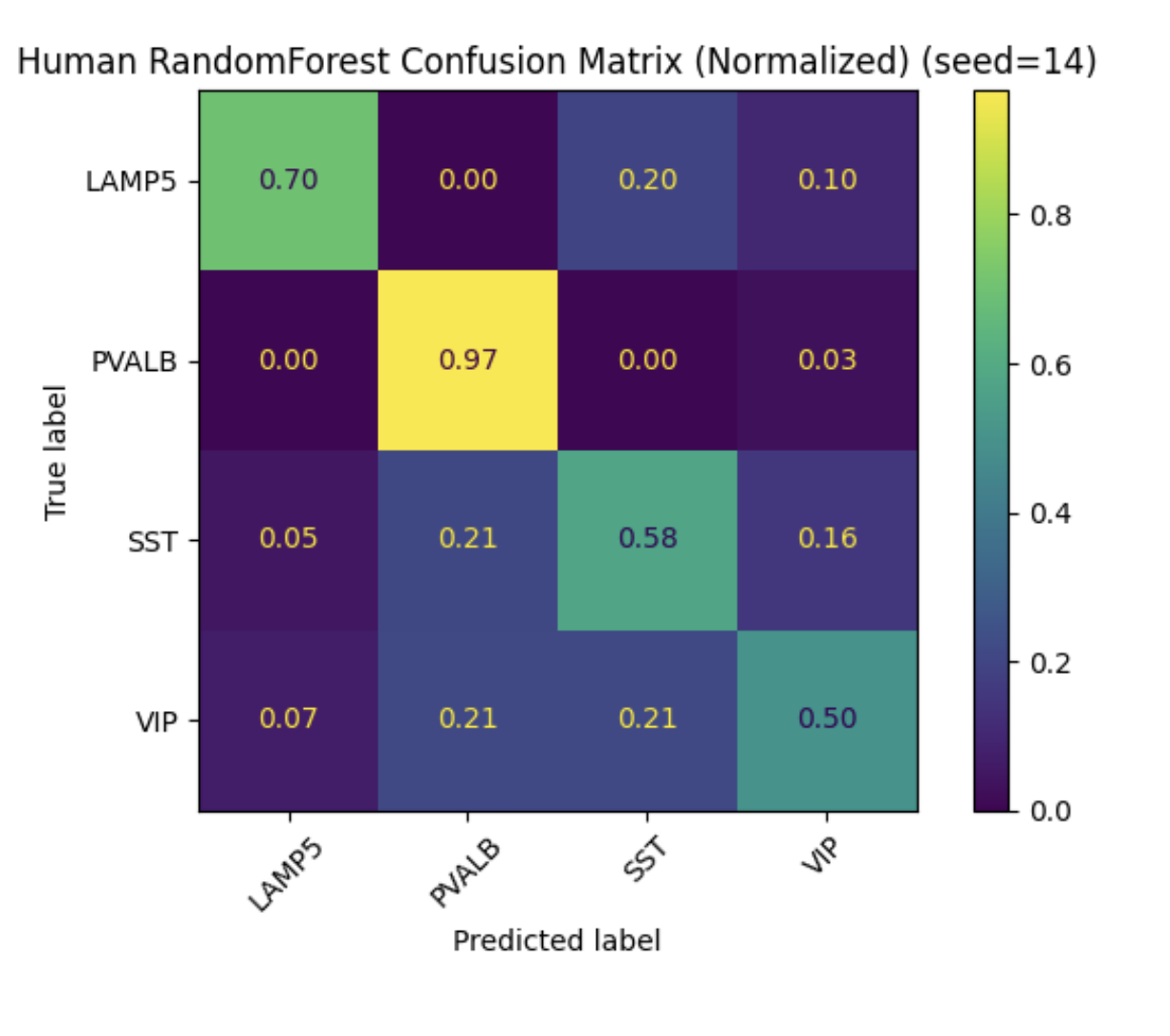} \\
\end{tabular}
\end{adjustbox}
\caption{Recap figure (reproduced for convenience). Representative confusion matrices (mouse left; human right) for the feature-based baseline classifier. These panels are repeated here as an at-a-glance summary of dominant error modes and class-imbalance effects; no new model fits are introduced in this figure.}
\label{fig:confusion-matrices}
\end{figure}

These limitations point to clear next steps. Increasing human sample size and focusing on well-matched inhibitory subclasses would allow a more rigorous test of whether separability improves with data alone or requires new feature representations. Incorporating morphology alongside electrophysiology is a natural extension. For ease of reference, the recap UMAP panel is shown in Figure~\ref{fig:umap-summary}, and the recap baseline confusion matrices are shown in Figure~\ref{fig:confusion-matrices}.

\FloatBarrier
\small
\bibliographystyle{plainnat}

\begin{thebibliography}{99}

\bibitem[Gouwens et~al.(2020)Gouwens, Sorensen, Baftizadeh, et~al.]{Gouwens2020} Gouwens, N. W., Sorensen, S. A., Baftizadeh, F., et al. (2020). Integrated morphoelectric and transcriptomic classification of cortical GABAergic cells. \textit{Cell}, 183(4), 935--953.e19. \url{https://doi.org/10.1016/j.cell.2020.09.057}

\bibitem[Allen Institute for Brain Science(2021)]{Allen2021} Allen Institute for Brain Science. (2021). Patch-seq recordings from mouse visual cortex (Version 0.210913.1639) [Dataset]. DANDI Archive. \url{https://doi.org/10.48324/dandi.000020/0.210913.1639}

\bibitem[Allen Institute for Brain Science(2023)]{Allen2023} Allen Institute for Brain Science. (2023). human\_patchseq\_gaba (metadata tables for Lee et al., 2023) [Data repository]. GitHub. \url{https://github.com/AllenInstitute/human_patchseq_gaba/tree/main/data} (accessed January 11, 2026)

\bibitem[Allen Institute for Brain Science(2024)]{Allen2024} Allen Institute for Brain Science. (2024). Human interneuron patch-seq electrophysiology (Version 0.241120.0510) [Dataset]. DANDI Archive. \url{https://doi.org/10.48324/dandi.000636/0.241120.0510}

\bibitem[Allen Institute for Brain Science(2025)]{AllenIPFX2025} Allen Institute for Brain Science. (2025). ipfx (Intrinsic Physiology Feature Extractor) (Version 2.1.2) [Computer software]. GitHub. \url{https://github.com/AllenInstitute/ipfx} (accessed January 11, 2026)

\bibitem[Bahdanau et~al.(2015)Bahdanau, Cho, and Bengio]{Bahdanau2015} Bahdanau, D., Cho, K., \& Bengio, Y. (2015). Neural machine translation by jointly learning to align and translate (arXiv:1409.0473). \url{https://doi.org/10.48550/arXiv.1409.0473}

\bibitem[Berg et~al.(2021)Berg, Sorensen, Ting, et~al.]{Berg2021} Berg, J., Sorensen, S. A., Ting, J. T., et al. (2021). Human neocortical expansion involves glutamatergic neuron diversification. \textit{Nature}, 598, 151--158. \url{https://doi.org/10.1038/s41586-021-03813-8}

\bibitem[Breiman(2001)]{Breiman2001} Breiman, L. (2001). Random forests. \textit{Machine Learning}, 45, 5--32. \url{https://doi.org/10.1023/A:1010933404324}

\bibitem[Cadwell et~al.(2016)Cadwell, Palasantza, Jiang, et~al.]{Cadwell2016} Cadwell, C. R., Palasantza, A., Jiang, X., et al. (2016). Electrophysiological, transcriptomic and morphologic profiling of single neurons using Patch-seq. \textit{Nature Biotechnology}, 34(2), 199--203. \url{https://doi.org/10.1038/nbt.3445}

\bibitem[Chartrand et~al.(2023)Chartrand et~al.]{Chartrand2023} Chartrand, T., et al. (2023). Morphoelectric and transcriptomic divergence of the layer 1 interneuron repertoire in human versus mouse neocortex. \textit{Science}. \url{https://doi.org/10.1126/science.adf0805}

\bibitem[Chawla et~al.(2002)Chawla, Bowyer, Hall, and Kegelmeyer]{Chawla2002} Chawla, N. V., Bowyer, K. W., Hall, L. O., \& Kegelmeyer, W. P. (2002). SMOTE: Synthetic minority over-sampling technique. \textit{Journal of Artificial Intelligence Research}, 16, 321--357. \url{https://doi.org/10.1613/jair.953}

\bibitem[Cui et~al.(2019)Cui, Jia, Lin, Song, and Belongie]{Cui2019} Cui, Y., Jia, M., Lin, T.-Y., Song, Y., \& Belongie, S. (2019). Class-balanced loss based on effective number of samples. In \textit{Proceedings of the IEEE/CVF Conference on Computer Vision and Pattern Recognition (CVPR)} (pp. 9268--9277).

\bibitem[Deng et~al.(2019)Deng, Guo, Xue, and Zafeiriou]{Deng2019} Deng, J., Guo, J., Xue, N., \& Zafeiriou, S. (2019). ArcFace: Additive angular margin loss for deep face recognition. In \textit{Proceedings of the IEEE/CVF Conference on Computer Vision and Pattern Recognition (CVPR)} (pp. 4690--4699). \url{https://doi.org/10.1109/CVPR.2019.00482}

\bibitem[Fuzik et~al.(2016)Fuzik, Zeisel, M\'at\'e, et~al.]{Fuzik2016} Fuzik, J., Zeisel, A., M\'at\'e, Z., et al. (2016). Integration of electrophysiological recordings with single-cell RNA-seq data identifies neuronal subtypes. \textit{Nature Biotechnology}, 34(2), 175--183. \url{https://doi.org/10.1038/nbt.3443}

\bibitem[Gouwens et~al.(2019)Gouwens, Sorensen, Baftizadeh, et~al.]{Gouwens2019} Gouwens, N. W., Sorensen, S. A., Baftizadeh, F., et al. (2019). Classification of electrophysiological and morphological neuron types in the mouse visual cortex. \textit{Nature Neuroscience}, 22(7), 1182--1195. \url{https://doi.org/10.1038/s41593-019-0417-0}

\bibitem[Hochreiter and Schmidhuber(1997)]{Hochreiter1997} Hochreiter, S., \& Schmidhuber, J. (1997). Long short-term memory. \textit{Neural Computation}, 9(8), 1735--1780. \url{https://doi.org/10.1162/neco.1997.9.8.1735}

\bibitem[Hodge et~al.(2019)Hodge, Bakken, Miller, et~al.]{Hodge2019} Hodge, R. D., Bakken, T. E., Miller, J. A., et al. (2019). Conserved cell types with divergent features in human versus mouse cortex. \textit{Nature}, 573, 61--68. \url{https://doi.org/10.1038/s41586-019-1506-7}

\bibitem[Kingma and Ba(2015)]{Kingma2015} Kingma, D. P., \& Ba, J. (2015). Adam: A method for stochastic optimization (arXiv:1412.6980). \url{https://doi.org/10.48550/arXiv.1412.6980}

\bibitem[Lee et~al.(2021)Lee, Budzillo, Hadley, et~al.]{Lee2021} Lee, B. R., Budzillo, A., Hadley, K., et al. (2021). Scaled, high fidelity electrophysiological, morphological, and transcriptomic cell characterization. \textit{eLife}, 10, e65482. \url{https://doi.org/10.7554/eLife.65482}

\bibitem[Lee et~al.(2023)Lee, Dalley, Miller, et~al.]{Lee2023} Lee, B. R., Dalley, R., Miller, J. A., et al. (2023). Signature morphoelectric properties of diverse GABAergic interneurons in the human neocortex. \textit{Science}, 382(6667), eadf6484. \url{https://doi.org/10.1126/science.adf6484}

\bibitem[Lin et~al.(2017)Lin, Goyal, Girshick, He, and Doll\'ar]{Lin2017} Lin, T.-Y., Goyal, P., Girshick, R., He, K., \& Doll\'ar, P. (2017). Focal loss for dense object detection. In \textit{Proceedings of the IEEE International Conference on Computer Vision (ICCV)} (pp. 2980--2988). \url{https://arxiv.org/abs/1708.02002}

\bibitem[Lipovsek et~al.(2021)Lipovsek, Bardy, Cadwell, et~al.]{Lipovsek2021} Lipovsek, M., Bardy, C., Cadwell, C. R., et al. (2021). Patch-seq: Past, present, and future. \textit{Journal of Neuroscience}, 41(5), 937--946. \url{https://doi.org/10.1523/JNEUROSCI.1653-20.2020}

\bibitem[Loshchilov and Hutter(2019)]{Loshchilov2019} Loshchilov, I., \& Hutter, F. (2019). Decoupled weight decay regularization. In \textit{International Conference on Learning Representations (ICLR)}. \url{https://arxiv.org/abs/1711.05101}

\bibitem[McInnes et~al.(2018)McInnes, Healy, Saul, and Gro\ss{}berger]{McInnes2018} McInnes, L., Healy, J., Saul, N., \& Gro\ss{}berger, L. (2018). UMAP: Uniform manifold approximation and projection. \textit{Journal of Open Source Software}, 3(29), 861. \url{https://doi.org/10.21105/joss.00861}

\bibitem[Nandi et~al.(2022)Nandi et~al.]{Nandi2022} Nandi, A., et al. (2022). Single-neuron models linking electrophysiology, morphology, and transcriptomics across cortical cell types. \textit{Cell Reports}, 41(6), 111659. \url{https://doi.org/10.1016/j.celrep.2022.111659}

\bibitem[R\"ubel et~al.(2022)R\"ubel, Tritt, Ly, et~al.]{Rubel2022} R\"ubel, O., Tritt, A., Ly, R., et al. (2022). The Neurodata Without Borders ecosystem for neurophysiological data science. \textit{eLife}, 11, e78362. \url{https://doi.org/10.7554/eLife.78362}

\bibitem[Tasic et~al.(2018)Tasic, Yao, Graybuck, et~al.]{Tasic2018} Tasic, B., Yao, Z., Graybuck, L. T., et al. (2018). Shared and distinct transcriptomic cell types across neocortical areas. \textit{Nature}, 563, 72--78. \url{https://doi.org/10.1038/s41586-018-0654-5}

\bibitem[Teeters et~al.(2015)Teeters et~al.]{Teeters2015} Teeters, J. L., et al. (2015). Neurodata Without Borders: Creating a common data format for neurophysiology. \textit{Neuron}, 88(4), 629--634. \url{https://doi.org/10.1016/j.neuron.2015.10.025}

\bibitem[Zou et~al.(2006)Zou, Hastie, and Tibshirani]{Zou2006} Zou, H., Hastie, T., \& Tibshirani, R. (2006). Sparse principal component analysis. \textit{Journal of Computational and Graphical Statistics}, 15(2), 265--286. \url{https://doi.org/10.1198/106186006X113430}

\end{thebibliography}


\end{document}